\documentclass[11pt]{article}

\usepackage[preprint]{acl}

\usepackage{times}
\usepackage{latexsym}
\usepackage[T1]{fontenc}
\usepackage[utf8]{inputenc}
\usepackage{microtype}
\usepackage{booktabs}
\usepackage{multirow}
\usepackage{xcolor,colortbl}
\usepackage{tcolorbox}
\usepackage{amssymb}
\usepackage{amsmath}
\usepackage{inconsolata}
\usepackage{graphicx}
\usepackage{arydshln}
\usepackage[table,dvipsnames]{xcolor}

\usepackage{xspace}
\newcommand{\duallaat}[1][]{\textsc{DualLAAT}\ifx#1\empty\else$_{\text{#1}}$\fi\xspace}

\usepackage{tikz}
\usetikzlibrary{arrows.meta, tikzmark}
\tikzset{improvearrow/.style={-{Triangle[length=1.5mm,width=1.4mm]}, line width=0.9pt, green!60!black}, shorten >=1pt, shorten <=2pt}

\newcommand{\uponly}[0]{\textcolor{ForestGreen}{\scriptsize($\uparrow$)}}
\newcommand{\up}[1]{\textcolor{ForestGreen}{\scriptsize(#1\,$\uparrow$)}}

\title{Bridging the Version Gap: Multi-version Training \\Improves ICD Code Prediction, Especially for Rare Codes}

\author{
   Jinghui Liu \hfill Anthony Nguyen \\
Australian e-Health Research Centre, CSIRO \\ 
  \texttt{\{jinghui.liu,anthony.nguyen\}@csiro.au} \\
}

\begin{document}
\maketitle
\begin{abstract}

Clinical coding maps clinical documentation to standardized medical codes, an essential yet time-consuming administrative task that could benefit from automation. Current models on ICD coding are typically optimized for codes from a specific ICD version. However, in reality, ICD systems evolve continuously, and different versions are adopted across time periods and regions. Moreover, ICD coding suffers from the long-tail problem, and rare code performance can be a bottleneck for developing implementable models. We examine whether it is viable to train version-independent models by combining data annotated in different ICD versions, which may help address these challenges. We add ICD-9 data to the training of a modified label-wise attention model for ICD-10 prediction, and find that despite the version mismatch, adding ICD-9 yields a 27\% increase in micro F1 for 18K rare ICD codes compared to training on ICD-10 alone. On 8K frequent ICD-10 codes, the multi-version training also substantially improves macro metrics, with far fewer model parameters.

\end{abstract}

%
\begin{figure}[t]
\centering
\begin{tikzpicture}[
    font=\scriptsize,
    >=Triangle,
    ds/.style    = {draw, rounded corners=2pt, minimum width=2.00cm, minimum height=0.50cm,
                    inner sep=2pt, font=\scriptsize, align=center},
    model/.style = {draw, rounded corners=3pt, fill=orange!15, minimum width=1.55cm,
                    minimum height=1.55cm, align=center, font=\scriptsize},
    panelHeader/.style = {font=\bfseries\footnotesize, anchor=west}
]

\begin{scope}[local bounding box=panela]
    \node[panelHeader] at (0, 2.45) {(a) Two challenges in ICD coding};

    \node[font=\scriptsize\itshape, anchor=west] at (0, 1.95) {ICD version evolution};

    \node[draw, rounded corners=2pt, fill=blue!12,  font=\scriptsize,
          minimum width=0.90cm, minimum height=0.42cm] (v9)  at (0.55, 1.35) {ICD-9};
    \node[draw, rounded corners=2pt, fill=green!12, font=\scriptsize,
          minimum width=0.90cm, minimum height=0.42cm] (v10) at (1.85, 1.35) {ICD-10};
    \node[draw, rounded corners=2pt, fill=gray!15,  font=\scriptsize,
          minimum width=0.90cm, minimum height=0.42cm] (v11) at (3.15, 1.35) {ICD-11};

    \draw[->]         (v9)  -- (v10);
    \draw[->, dashed] (v10) -- (v11);

    \node[font=\tiny, anchor=north] at (1.20, 1.10) {major shift};
    \node[font=\tiny, anchor=north] at (2.50, 1.10) {upcoming};

    \node[font=\scriptsize, anchor=north west, text width=3.55cm, align=left, inner sep=1pt]
         at (0.1, 0.58)
         {\itshape Models trained for one version do not transfer to others.};

    \node[font=\scriptsize\itshape, anchor=west] at (3.85, 1.95) {Long-tail code distribution};

    \begin{scope}[xshift=3.95cm, yshift=0.40cm]
        \foreach \x/\h in {0/1.20, 0.13/0.96, 0.26/0.74, 0.39/0.55, 0.52/0.40, 0.65/0.29} {
            \draw[fill=blue!40, draw=blue!70, line width=0.3pt] (\x, 0) rectangle ++(0.10, \h);
        }
        \foreach \x/\h in {0.78/0.20, 0.91/0.15, 1.04/0.11, 1.17/0.08, 1.30/0.06,
                           1.43/0.05, 1.56/0.04, 1.69/0.04, 1.82/0.03, 1.95/0.03,
                           2.08/0.03, 2.21/0.02, 2.34/0.02, 2.47/0.02, 2.60/0.02,
                           2.73/0.02, 2.86/0.02} {
            \draw[fill=red!55, draw=red!75, line width=0.3pt] (\x, 0) rectangle ++(0.10, \h);
        }
        \draw[->, thin] (-0.05, 0) -- (3.12, 0);
        \node[font=\tiny, blue!55!black, anchor=west] at (0.20, 1.05) {8K freq.};
        \node[font=\tiny, red!70!black,  anchor=west] at (1.00, 0.45) {18K rare ($\sim$70\%)};
        \node[font=\scriptsize, anchor=north, text width=3.35cm, align=left, inner sep=1pt] at (1.55, -0.04) {\textit{Low-frequency codes (<10) predominate.}};
    \end{scope}
\end{scope}

\begin{scope}[yshift=-3.10cm, local bounding box=panelb]
    \node[panelHeader] at (0, 2.55) {(b) Mixed-version training};

    \node[ds, fill=blue!12,  anchor=west] (d1) at (0, 1.95) {MIMIC-III \textit{ICD-9}};
    \node[ds, fill=blue!12,  anchor=west] (d2) at (0, 1.25) {MIMIC-IV \textit{ICD-9}};
    \node[ds, fill=green!15, anchor=west] (d3) at (0, 0.55) {MIMIC-IV \textit{ICD-10}};

    \node[model] (m) at (3.85, 1.25)
        {\textsc{DualLAAT}\\[2pt]
         \scriptsize adaptively encode \\
         \scriptsize notes and codes\\[-1pt]
         \scriptsize {\itshape w/} LWA};

    \draw[->] (d1.east) -- (m.west |- d1.east);
    \draw[->] (d2.east) -- (m.west);
    \draw[->] (d3.east) -- (m.west |- d3.east);

    \node[draw, rounded corners=3pt, fill=yellow!25, font=\scriptsize, align=center,
          minimum width=1.80cm, minimum height=1.55cm] (out) at (6.35, 1.25)
          {Version-agnostic\\ ICD predictions};
    \draw[->] (m.east) -- (out.west);
\end{scope}

\begin{scope}[yshift=-5.85cm, local bounding box=panelc]
    \node[panelHeader] at (0, 2.30) {(c) Gains on ICD-10 (adding ICD-9 to training)};

    \node[font=\scriptsize\itshape, anchor=west] at (0, 1.80) {18K rare -- Micro F1};

    \node[font=\scriptsize, anchor=east] at (1.05, 1.30) {ICD-10};
    \draw[fill=gray!45, draw=gray!75]               (1.10, 1.15) rectangle (1.94, 1.45);
    \node[font=\scriptsize, anchor=west] at (1.99, 1.30) {8.4};

    \node[font=\scriptsize, anchor=east] at (1.05, 0.70) {$+\,$ICD-9};
    \draw[fill=ForestGreen!60, draw=ForestGreen!80] (1.10, 0.55) rectangle (2.17, 0.85);
    \node[font=\scriptsize, anchor=west] at (2.22, 0.70)
        {10.7\, \textcolor{ForestGreen}{\textbf{(+27\%)}}};

    \node[font=\scriptsize\itshape, anchor=west] at (4.00, 1.80) {8K frequent -- Macro F1};

    \node[font=\scriptsize, anchor=east] at (5.05, 1.30) {ICD-10};
    \draw[fill=gray!45, draw=gray!75]               (5.10, 1.15) rectangle (6.04, 1.45);
    \node[font=\scriptsize, anchor=west] at (6.09, 1.30) {27.9};

    \node[font=\scriptsize, anchor=east] at (5.05, 0.70) {$+\,$ICD-9};
    \draw[fill=ForestGreen!60, draw=ForestGreen!80] (5.10, 0.55) rectangle (6.14, 0.85);
    \node[font=\scriptsize, anchor=west] at (6.19, 0.70)
        {29.9\, \textcolor{ForestGreen}{\textbf{(+7\%)}}};
\end{scope}

\end{tikzpicture}

\caption{
\textbf{(a)} ICD coding faces two intertwined challenges: the ICD system evolves continuously
and the code distribution is heavily long-tailed.
\textbf{(b)} We mix three MIMIC-derived datasets spanning ICD-9 and ICD-10 to train a single version-agnostic model, \duallaat[].
\textbf{(c)} Adding ICD-9 to ICD-10 training yields a 27\% relative gain in micro F1 on rare ICD-10 codes than training on ICD-10 alone.
}
\label{fig:intro}
\end{figure}

\section{Introduction}

The International Classification of Diseases (ICD) is the global standard for reporting health conditions and diseases. It plays a critical role in healthcare billing, epidemiological research, and health policymaking~\cite{Dong2022-ly,Gan2025-ge}.
Assigning ICD codes to clinical documents requires a high level of expertise given that the candidate code set is often large (e.g., ICD-10 includes over 70K diagnosis codes) and the documentation can be lengthy and complex~\cite{Motzfeldt2025-pl,Liu2023-hp}. Consequently, the coding process is time-consuming even for experienced coders, which has motivated extensive research into NLP models that automate or assist the coding process~\cite{Stanfill2010-lz,Ji2024-ca}.

Current best-performing ICD coding models typically frame the task as a multi-label, multi-class classification problem~\cite{Douglas2025-jg, Edin2023-zj, Yuan2022-qx}, which requires defining a fixed label space and a substantial amount of data for supervised training.
This setup faces two critical challenges in real-world ICD coding. First, \textbf{the ICD coding system undergoes continuous updates}, with new codes added and old codes retired. For example, CMS actively updates ICD-10 every year to keep codes aligned with healthcare needs.\footnote{https://www.cms.gov/medicare/coding-billing/icd-10-codes} At certain points, the ICD system undergoes major transitions, such as the shift from ICD-9 to ICD-10, and the anticipated move to ICD-11~\cite{Harrison2021-hf}.
Models with a pre-defined label space have trouble handling these updates, which introduce substantial data shifts~\cite{Finlayson2021-re}.

Second, \textbf{ICD code distributions are long-tailed}~\cite{Li2025-fa,He2025-wm}, which is a critical modeling challenge as the ICD system grows larger and more fine-grained. For example, in MIMIC-IV~\cite{Johnson2023-qk}, infrequent ICD-10 codes that appear fewer than 10 times (18K) account for 69.6\% of all codes in the database (26K). Collecting sufficient samples for these rare and underrepresented diseases is essential for training.
However, achieving this requires consolidating patient records
across time periods and regions, which is likely accompanied by mismatched ICD versions.
It remains understudied whether such a combination leads to benefits or harm in modeling.
For example, given that only 24.3\% of ICD-9-CM codes have exact matches in ICD-10-CM~\cite{Fung2021-ye}, mixing multiple datasets across versions may plausibly introduce noise that harms rather than helps modeling.

This study explores this question by examining if ICD-9 data is valuable for training an ICD-10 prediction model (Figure~\ref{fig:intro}).
Our experiments used two ICD-9 datasets and one ICD-10 dataset based on MIMIC~\cite{Edin2023-zj}, and mixed their training sets to train a version-independent model based on label-wise attention (LWA)~\cite{Wu2024-lb,Mullenbach2018-zn}.
Adding ICD-9 data to the training set led to
\textbf{substantially improved performance on rare codes} (micro F1: $8.4\rightarrow10.7$) in ICD-10
using \textbf{a much smaller model},
as well as
\textbf{improved per-label performance on frequent codes}  (macro F1 of 29.9).
Our findings demonstrate the synergistic effect of combining multi-version datasets for ICD coding,
showcasing the potential value of leveraging legacy patient records to train more robust ICD prediction systems.

\section{Related Work}

Research on automated and assisted ICD coding dates back to the 1990s~\cite{Larkey1996-zl}.
Recent advances have largely focused on configuring neural architectures to better model the large code set~\cite{Li2025-ub,Shi2017-qw}.
\citet{Ji2024-ca} surveyed modeling approaches and highlighted that label-wise attention (LWA)~\cite{Mullenbach2018-zn,Wu2024-lb,Vu2020-em} remains a key module to train high-performing models. \citet{Edin2023-zj} conducted a comprehensive benchmark of various LWA-based models across three datasets spanning MIMIC-III and IV and ICD-9 and ICD-10, finding that PLM-ICD with in-domain pretrained checkpoints performed best.
Other approaches exploit the relations between codes in the ICD ontology to improve modeling~\cite{Luo2024-ih,Yuan2022-qx}.
To the best of our knowledge, all prior models were trained for a specific set of ICD codes as label space, and none were evaluated across different ICD versions.

A growing number of studies have explored prompting LLMs to predict ICD codes~\cite{Simmons2025-ta,Yang2023-xb}. While simple prompting achieves low performance~\cite{Soroush_Ali2024-vw}, agent-based systems for ICD prediction show promise~\cite{Motzfeldt2025-pl,Zheng2025-tm}. The generative approach is not constrained by a fixed label space and may offer enhanced interpretability. However, most of these studies were conducted on small-scale datasets with a limited label space of about 1K~\cite{Cheng2023-ag}, which contrasts with existing benchmarks (8K in \citet{Edin2023-zj}) and public databases (26K in \citet{Johnson2023-qk}). In addition, agent-based approaches often require large token budgets, which can be costly.
For this study, we focus on comparing with supervised approaches on datasets with large code sets.

\section{Methods and Experiments}
\subsection{Benchmark Datasets}

We adopted the benchmark from \citet{Edin2023-zj}, which includes three MIMIC-derived datasets, namely MIMIC-III (ICD-9), MIMIC-IV (ICD-9), and MIMIC-IV (ICD-10).
The original benchmark focuses on frequent codes, defined as appearing $\ge10$ times in the corresponding database.
To also evaluate rare codes, we re-extracted all codes from MIMIC, resulting in substantial expansions of the code space ($6K\rightarrow11K$ for ICD-9 and $8K\rightarrow26K$ for ICD-10).
Since ICD-9 is no longer in active use, we focused on ICD-10 as the primary evaluation benchmark.
Dataset statistics are presented in Appendix Table~\ref{tab:datasets}.

\subsection{Classification Model}
LWA trains code-specific embeddings to attend to relevant text tokens in clinical notes and uses them for classification~\cite{Edin2024-vm}. To enable adaptive modeling across ICD versions, we trained an encoder to generate code embeddings from code descriptions, which are unique textual strings describing each ICD code.
We jointly trained two text encoders -- one for clinical notes and one for ICD codes -- using LWA. Given the dual-encoder feature and the use of LWA from LAAT~\cite{Vu2020-em}, we denote the model as \duallaat[].
The model takes $N$ notes and $C$ codes (as text strings) as input and outputs a probability matrix $\hat{\mathbf{y}}\in\mathbb{R}^{N \times C}$ as prediction. The codes in $C$ are version-agnostic and can be ICD-9 or ICD-10. Detailed notations and training procedures
are provided in Appendix~\ref{sec:model}.

\begin{table*}[t]
    \centering
    \resizebox{\textwidth}{!}{
    \begin{tabular}{lcccccccc}
        \toprule
        & \multicolumn{4}{c}{Classification} & \multicolumn{4}{c}{Ranking} \\
        \cmidrule(lr){2-5}\cmidrule(lr){6-9}
         & \multicolumn{2}{c}{AUC-ROC} & \multicolumn{2}{c}{F1} & \multicolumn{2}{c}{Precision@k} & R-precision & MAP \\
         & Micro & Macro & Micro & Macro & 8 & 15 &  &  \\
        \midrule
        \multicolumn{9}{c}{\textit{MIMIC-III ICD-9}} \\
        PLM-ICD~\cite{Edin2023-zj} & 98.9 & 95.9 & 59.6 & 26.6 & 72.1 & 56.5 & 60.1 & 64.6 \\
        \hdashline[0.8pt/4pt]
        \duallaat  & \bfseries  99.2 & \bfseries  96.9 & \bfseries  61.8 & \bfseries  33.3 & \bfseries  74.9 & \bfseries  58.8 & \bfseries  62.8 & \bfseries  68.1 \\
        \midrule
        \multicolumn{9}{c}{\textit{MIMIC-IV ICD-9}} \\
        PLM-ICD~\cite{Edin2023-zj}  & 99.4 & 97.2 & 62.6 & 29.8 & 70.0 & 53.5 & 62.7 & 68.0 \\
        \hdashline[0.8pt/4pt]
        \duallaat  & \bfseries  99.5 & \bfseries  97.6 & \bfseries  63.4 & \bfseries  34.3 & \bfseries  71.0 & \bfseries  54.3 & \bfseries  63.8 & \bfseries  69.3 \\
        \midrule
        \multicolumn{9}{c}{\textit{MIMIC-IV ICD-10}} \\
        PLM-ICD~\cite{Edin2023-zj}   & 99.2 & 96.6 & \bfseries 58.5 & 21.1 & \bfseries 69.9 & \bfseries 55.0 & 57.9 & 61.9 \\
        \hdashline[0.8pt/4pt]
        \duallaat[cnn] (ICD-10 only) & 99.2 & 96.8 & 55.6 & 24.3 & 67.5 & 52.7 & 55.3 & 59.0 \\
        \duallaat[cnn]  &  99.3 \uponly & 97.2 \uponly & 56.6 \uponly & 26.0 \uponly & 68.4 \uponly & 53.6 \uponly & 56.4 \uponly & 60.4 \uponly \\
        \hdashline[0.8pt/4pt]
        \duallaat[] \ \ \ \   (ICD-10 only) & 99.3 & 97.1  & 57.5 & 27.9 & 69.3 & 54.4 & 57.2 & 61.5 \\
        \duallaat[] &  \bfseries 99.3 \textcolor{ForestGreen}{\scriptsize($-$)} &   \bfseries 97.4 \uponly & 58.0 \uponly &  \bfseries 29.9 \uponly & \bfseries 69.9 \uponly & 54.9 \uponly &  \bfseries 58.0 \uponly &  \bfseries 62.3 \uponly \\
        \bottomrule
    \end{tabular}}
    \caption{
    Results on frequent ICD codes for MIMIC-III \textit{ICD-9} (3.7K), MIMIC-IV \textit{ICD-9} (6K) and MIMIC-IV  \textit{ICD-10} (8K).
    The means of three random seed runs were reported.
    For ICD-10 prediction,
    \textcolor{ForestGreen}{$\uparrow$} indicates improvement of multi-source training (ICD-10 + ICD-9) compared to training on ICD-10 alone.
    }
    \label{tab:main_results}
\end{table*}

\subsection{Experiments}

To investigate the impact of mixing multiple ICD data sources, we aggregated the three train sets from MIMIC-III (ICD-9), MIMIC-IV (ICD-9), and MIMIC-IV (ICD-10) for training and evaluated on the corresponding test sets.
We included all ICD codes (i.e., from full label space) in the aggregated data to train the model.
To enable fast experimentation and to highlight the impact of data mixing, we used basic CNN~\cite{Mullenbach2018-zn} and RNN~\cite{Vu2020-em} as text encoders, denoted as \duallaat[cnn] and \duallaat, respectively.

We adopted the benchmark metrics~\cite{Edin2023-zj} for evaluation on frequent codes, including both classification and ranking metrics.
In particular, macro F1 was calculated as the arithmetic mean of F1 scores per code.
Higher macro F1 indicates stronger performance on less frequent codes, reflecting tail-end robustness.
We also evaluated on full ICD-10 set and explicitly on rare codes, which in practice can be equally important as frequent codes for reporting disease prevalence and enabling appropriate reimbursement.

\subsection{Baselines}
We consider two strong supervised models as baselines for frequent and full code sets, respectively:

\textbf{PLM-ICD.} This model adopts in-domain BERT with sliding-window as the text encoder and trains with LWA~\cite{Huang2022-fn}.
Prior work found replacing BERT with Llama on a dataset with 50 codes did not improve performance~\cite{Motzfeldt2025-pl}, highlighting the strong baseline performance of PLM-ICD.
We report its results from the previous benchmark study on the frequent 8K codes~\cite{Edin2023-zj}.

\textbf{CoRelation.} This model improves code representation learning by modeling the relations between codes through a bipartite graph for enhanced accuracy and efficiency~\cite{Luo2024-ih}. It was evaluated on top-50 and full codes in MIMIC.
We use this model for reference comparison on full ICD-10 set with 26K codes.

\section{Results}

\subsection{Impact of Data Mixing on Frequent Code Prediction and Benchmark Comparison}

Table~\ref{tab:main_results} presents the results on frequent codes. On ICD-10 prediction, mixing additional ICD-9 datasets from MIMIC improves all metrics for both \duallaat[cnn] and \duallaat[rnn], with about 2 points of improvement on macro F1.
This shows that adding ICD-9 to training is beneficial for ICD-10 prediction despite terminological differences.
Using a stronger encoder (RNN over CNN) also brings steady improvements, consistent with prior findings~\cite{Edin2023-zj} and suggesting potential gains with stronger encoder models.

Compared with the PLM-ICD baseline, RNN-based \duallaat[] trained on mixed MIMIC datasets achieved similar precision but much higher macro AUC-ROC and macro F1.
We also report results for the weaker CAML~\cite{Mullenbach2018-zn} and LAAT~\cite{Vu2020-em} in Appendix Table~\ref{tab:std}. These baseline methods with CNN and RNN encoders form a more direct comparison with our model. The additional results also show that \duallaat trained using mixed datasets offers lower variance across the metrics.

\subsection{Impact on Rare Code Prediction}

Mixing additional data sources shows substantial benefits in rare code prediction, as presented in Table~\ref{tab:rare_results}. Both ICD-10 and 9 benefited from alternative ICD data. For ICD-10, this yielded a 27.4\% increase in micro F1, which is a substantial improvement given the number of codes (18K).

Table~\ref{tab:full} compares the  model performance on the full code set with 26K labels from MIMIC-IV ICD-10. \duallaat achieves results comparable to the reported CoRelation numbers on shared metrics, with a notably higher macro F1 (6.3$\rightarrow$11.2), suggesting stronger performance across the long tail of rare codes.
This shows that enhancing training data can be as important in ICD coding as architectural improvements, which have been the predominant focus in ICD prediction literature~\cite{Ji2024-ca,Dong2022-ly}.

\subsection{Training and Parameter Efficiency}

Mixing multiple ICD sources for training also improves efficiency as the model can now be applied to datasets with different ICD versions, in effect reducing the number of model parameters required.
In the context of this study, three PLM-ICD models are needed for the three benchmarks on frequent code prediction, whereas one \duallaat can handle them all.
Training with mixed data also shortens convergence time: our training plateaued at around 10 epochs compared to 20 epochs for PLM-ICD.

Table~\ref{tab:efficiency} reports the comparison between PLM-ICD and \duallaat. While \duallaat[] with mixed datasets achieves comparable performance, it uses only a fraction of the training time and parameters. This can be an important factor to consider when deploying a model in a low-resource hospital or outpatient setting~\cite{Wu2022-pg}.

\begin{table}[h!]
    \centering
    \resizebox{\linewidth}{!}{
\begin{tabular}{lccc}
    \toprule
    & F1 Micro & Precision@8 & MAP \\
    \midrule
    \multicolumn{4}{c}{\textit{MIMIC-IV-ICD10}} \\
    Single Dataset & 8.4 & 5.8 & 19.5 \\
    \ + Mixing Alt ICD & 10.7 \up{27.4\%} & 6.1 \up{5.2\%} & 21.4 \up{9.7\%} \\
    \multicolumn{4}{c}{\textit{MIMIC-IV-ICD9}} \\
    Single Dataset & 7.2 & 5.6 & 25.4 \\
    \ + Mixing Alt ICD & 10.9 \up{51.4\%} & 6.6 \up{17.9\%} & 31.2 \up{22.8\%} \\
    \bottomrule
\end{tabular}
}
\caption{Results on the rare codes (frequency < 10 in the cohort).
\duallaat[cnn] trained on the corresponding train set was used as baseline.
}
    \label{tab:rare_results}
\end{table}

\begin{table}[]
    \centering
    \resizebox{0.9\linewidth}{!}{
    \begin{tabular}{lcc}
    \toprule
         &  Training time & \# Param \\
         \midrule
    \multicolumn{3}{c}{\textit{ICD-10 only}} \\
      PLM-ICD    &  34 hrs & 137M \\
      \duallaat[cnn] & 5 hrs & 15M \\
      \duallaat[] & 18 hrs & 37M \\
      
     \midrule
    \multicolumn{3}{c}{\textit{ICD-10 + ICD-9 ($\times3$ datasets)} } \\
      PLM-ICD ($\times3$ models)   &  95 hrs & 402M \\
      \duallaat[cnn] & 17 hrs & 15M \\
      \duallaat[] & 57 hrs & 37M \\
      \bottomrule
    \end{tabular}
    }
    \caption{Comparison of training time and model size. Mixed training data includes one dataset for ICD-10 and two datasets for ICD-9.}
    \label{tab:efficiency}
\end{table}

\begin{table*}[h!]
\centering
    \resizebox{\textwidth}{!}{
\begin{tabular}{lccccc}
    \toprule
    & AUC-ROC Micro & AUC-ROC Macro & F1 Micro & F1 Macro & Precision@8 \\
    \midrule
    PLM-ICD~\cite{Luo2024-ih}    & 99.0 & 91.9 & 57.0 & 4.9 & 69.5 \\
    CoRelation~\cite{Luo2024-ih}                   & 99.6          & 97.2          & 57.8 & 6.3           &  70.0  \\
    \duallaat (ICD-10 only)  & 99.6          & 97.0          & 56.6          & 10.2          & 69.3          \\
    \duallaat      & 99.7 & 97.4 & 57.1          & 11.2 & 69.8          \\
    \bottomrule
\end{tabular}
}
\caption{Results on 26K ICD-10 codes in MIMIC-IV. PLM-ICD and CoRelation results are from the original paper, which uses a different preprocessing pipeline; we include this approximate comparison for reference.}
\label{tab:full}
\end{table*}

\section{Discussion \& Conclusion}

Training effective ICD coding models is challenging due to the long-tail distribution of codes and the continuous updates to the ICD system. Most existing work focuses on improving modeling methods, yet little attention has been paid to the impact of the training data itself. 
We show that mixing multiple data sources -- even with varied ICD versions -- benefits both frequent and rare code predictions.
The gains are notable given that less than 25\% of ICD-9 have equivalent mappings to ICD-10~\cite{Fung2021-ye}, suggesting the benefits arise from shared clinical semantics rather than direct label overlap, which warrants future investigation.

The viability of mixing ICD versions means \textbf{legacy patient records can be utilized in training new models}, an important implication for developing coding systems to handle ongoing modifications of ICD rules. For major rule updates, a model trained on legacy records can avoid cold-start issues and serve as a foundation for fine-tuning.

In conclusion, we demonstrate the advantages of mixing multi-source datasets to train ICD prediction models, resulting in more accurate, robust, and efficient performance. The synergistic effect of combining diverse sources demonstrates the potential for further scaling both training data and model capacity to advance the progress of ICD coding research. Finally, our code and model checkpoints are released to support reproducibility.\footnote{The training code and model checkpoints can be found in \url{https://github.com/JHLiu7/Dual-LAAT}.}

\section*{Limitations}

\textbf{ICD and Data Coverage.} This study only considered the MIMIC databases as the data sources, which originate from a single institution in the United States and use ICD Clinical Modification (CM). Other regional implementations of ICD exist, such as ICD-10-CA for Canada and ICD-10-AM for Australia. Furthermore, many countries adopt ICD systems in non-English languages, including China (CCD/ICD-10-CN) and Germany (ICD-10-GM). We were limited by the scope of data in this initial study and focused on openly available datasets. Given insights from multilingual language modeling,  cross-lingual knowledge transfer~\cite{Artetxe2020-zp} may also be feasible for ICD modeling. However, this warrants future research, particularly to address nuanced differences between different ICD versions beyond mere translation and semantic mappings.

\textbf{Classification Model.} For this study, we employed simple classification models in the experiments, but we believe the patterns of data mixing can be extrapolated to other models given the existing findings on modeling architectures~\cite{Ji2024-ca,Edin2023-zj}. Training PLM-based models with multi-source datasets represents a natural follow-up, though we leave this open as newer and more effective modeling methods may emerge in the near future.

\textbf{Record Overlap.} An inherent challenge in mixing data sources is potential overlapping records. MIMIC-III (2001-2012) and MIMIC-IV (2008-2018) overlap temporally, but we could not explicitly identify shared patients because they use different ID conventions and are not linkable. This may have influenced results for ICD-9, although these were not the primary focus of our experiments.
Meanwhile, this is less of an issue for ICD-10
since it is distinctly different from ICD-9 and only exists in MIMIC-IV. We found seven patients appearing in both ICD-9 and ICD-10 datasets from \citet{Edin2023-zj} in MIMIC-IV, with only one patient in the ICD-10 test set (19,802 patients). Given the large test size and different data distributions, we deemed it acceptable to retain this patient to enable direct comparison with the baselines.

\bibliography{custom}

\appendix

\begin{table*}[ht!]
    \centering

    \resizebox{0.8\textwidth}{!}{
    \begin{tabular}{lccc}
        \toprule
        & {MIMIC-III \textit{ICD-9}}  & {MIMIC-IV \textit{ICD-9}}  & {MIMIC-IV \textit{ICD-10}}\\
        \midrule
        Number of notes  & 52,712 & 209,326 & 122,278 \\
        Notes: Train/val/test [\%] & 72.9/10.6/16.6  & 73.8/10.5/15.7& 72.9/10.9/16.2  \\
        \hline

        \multicolumn{4}{c}{\textit{Frequent Code Set}} \\
        Number of unique codes &  3,681 & 6,150 & 7,942  \\
Codes per note: Median (IQR) &  14 (10-20) & 12 (8-17) & 14 (9-20)  \\
Codes per note: Mean (Std) &  15.6$\pm$8.0 & 13.3$\pm$7.6 & 15.7$\pm$8.7  \\
        \hline

        \multicolumn{4}{c}{\textit{Full Code Set}} \\
        Number of unique codes &  8,925 & 11,324 & 26,085  \\
Codes per note: Median (IQR) &  14 (10-20) & 12 (8-17) & 15 (10-21)  \\
Codes per note: Mean (Std) &  15.9$\pm$8.1 & 13.4$\pm$7.6 & 16.0$\pm$8.9  \\
        \hline

        \multicolumn{4}{c}{\textit{Rare Code Set}} \\
Number of unique codes &  5,244 & 5,174 & 18,143  \\
Codes per note: Median (IQR) &  1 (1-2) & 1 (1-1) & 1 (1-2)  \\
Codes per note: Mean (Std) &  1.5$\pm$1.1 & 1.2$\pm$0.6 & 1.6$\pm$1.2  \\
Number of notes (at least 1 rare code) &  10,653 & 14,185 & 30,332  \\

        \bottomrule
    \end{tabular}
    }

    \caption{Benchmark datasets on MIMIC-III (ICD-9), MIMIC-IV (ICD-9), and MIMIC-IV (ICD-10) from \citet{Edin2023-zj}, which are based on the \textit{frequent code set}. Additional \textit{full code set} and \textit{rare code set} are also examined. Rare codes refer to codes occurring less than 10 times in the dataset, and were not included in \citet{Edin2023-zj}. \textit{Full code set} share the same clinical notes with the \textit{frequent code set}. IQR: interquartile range. Std: standard deviation. }
    \label{tab:datasets}

\end{table*}

\section{Datasets for Evaluation}
\label{sec:data}

Table~\ref{tab:datasets} presents the statistics of the three benchmark datasets from \citet{Edin2023-zj}. We additionally extracted the infrequent codes that were not included in the prior study to evaluate the models' robustness on tail-end ICD codes.

\section{Classification Model in Detail}
\label{sec:model}

\subsection{\duallaat with dual encoders}

Given a clinical note $N$ with $t_{\text{note}}$ word tokens $w^N_1, w^N_2, \ldots, w^N_{t_{\text{note}}}$, we have a clinical code $C$ with $t_{\text{code}}$ word tokens $w^C_1, w^C_2, \ldots, w^C_{t_{\text{code}}}$ from the code description. They are transformed into word embeddings by a shared embedding layer into $\mathbf{e}^N_1, \mathbf{e}^N_2, \ldots, \mathbf{e}^N_{t_{\text{note}}}$ and $\mathbf{e}^C_1, \mathbf{e}^C_2, \ldots, \mathbf{e}^C_{t_{\text{code}}}$, respectively.

Clinical note representations at token level $\mathbf{H}_{\text{note}} \in \mathbb{R}^{d_{\text{note}} \times t_{\text{note}}}$ are created by a note encoder as $encoder_{note}$:

\begin{equation}\label{eq:encode}
    \mathbf{H}_{\text{note}} = \text{encoder}_{\text{note}}(\mathbf{e}^N_1, \ldots, \mathbf{e}^N_{t_{\text{note}}})
\end{equation}

The clinical code representation is encoded by a separate
$encoder_{code}$
and pooled into a code-level representation. The representation for one code $\mathbf{h}_{\text{code}} \in \mathbb{R}^{d_{\text{code}}}$ is created as:

\begin{equation}
    \mathbf{h}_{\text{code}} = \text{Pooling}(\text{encoder}_{\text{code}}(\mathbf{e}^C_1, \ldots, \mathbf{e}^C_{t_{\text{code}}}))
\end{equation}

In practice, we consider $L$ codes by encoding them in a batch to create $\mathbf{H}_{\text{code}} \in \mathbb{R}^{|L| \times d_{\text{code}}}$. Then we have two projection matrices $\mathbf{W}_{\text{note}} \in \mathbb{R}^{d_{\text{shared}} \times d_{\text{note}}}$ and $\mathbf{W}_{\text{code}} \in \mathbb{R}^{d_{\text{code}} \times d_{\text{shared}}}$ for note and code, respectively.
The projected code representations $\mathbf{H}_{\text{code}} \cdot \mathbf{W}_{\text{code}} \in \mathbb{R}^{|L| \times d_{\text{shared}}}$
replaces the code-specific embeddings in typical LWA implementation.
With an additional dimension $d_{\text{shared}}$ specified,
the attention scores and representations based on dual encoders are computed:

\begin{align}
    \mathbf{A}_{\text{dual}} &= \text{softmax}\Big( \tanh(\mathbf{H}_{\text{code}} \cdot \mathbf{W}_{\text{code}}) \\
    &\qquad\qquad \cdot \tanh(\mathbf{W}_{\text{note}} \cdot \mathbf{H}_{\text{note}}) \Big) \nonumber \\
    \mathbf{J}_{\text{dual}} &= \mathbf{H}_{\text{note}} \cdot \mathbf{A}_{\text{dual}}^{\top}
\end{align}

This formulates the backbone of \duallaat, which use separate note encoder and code encoder to create their corresponding representations and compute label-wise attention.

To further enhance the dual label-wise attention, we also consider $M$ attention heads with $\mathbf{W}_{\text{note}}^1, \ldots, \mathbf{W}_{\text{note}}^M$ and $\mathbf{W}_{\text{code}}^1, \ldots, \mathbf{W}_{\text{code}}^M$ as multi-head attention (MHA)~\cite{Vaswani2017-cu}. Then we compute $\mathbf{J}_{\text{dual}}^1, \ldots, \mathbf{J}_{\text{dual}}^M$ and obtain the multi-head note representation per code by concatenation:

\begin{equation}
    \mathbf{J}_{\text{dual}}^{\text{mha}} = \text{concat}(\mathbf{J}_{\text{dual}}^1, \ldots, \mathbf{J}_{\text{dual}}^M)
\end{equation}

The final output is computed as:

\begin{equation}
    \hat{\mathbf{y}} = \text{sigmoid}(\text{classifier}(\mathbf{J}_{\text{dual}}^{\text{mha}}))
\end{equation}

The model is trained to minimize binary cross entropy loss. Different from the previous models that consider only note $N$ as input, \duallaat now takes both $N$ (notes) and $C$ (codes) as inputs. This enables a flexible choice of $L$ by changing
$C$  either during training or at inference. Since the clinical codes $C$ are provided as code descriptions in text,
it allows flexibility to feed different versions of ICD codes as long as they are provided in the textual form. During inference, these descriptions can be directly mapped back to the actual codes.

\subsection{Training \duallaat}

To train \duallaat in batches, we need to dynamically set the label space $L$ for different codes.
For example, given a batch of 32 patients with each patient having 20 codes on average, $L$ would be set dynamically according to appearing clinical codes with $|L|=640$  (20$\times$32).
Meanwhile, in reality the number of clinical codes can  vary dramatically from patient to patient, resulting in large variations in code numbers that would  destabilize the label space and thus training. In addition, patients in a batch sometimes may contain codes in overlap, in effective reducing the scope of $L$.

To address this, we specify the label space $L$ to be a fixed number that is larger than the possible codes to appear in a batch. Given the fixed $L$,  we collect the unique codes in the batch as positives $L_{pos}$, and then randomly sample negative codes $L_{neg}$ from a code pool to form $L$ so that $|L|=|L_{pos}|+|L_{neg}|$. This setup has three advantages. First, it fixes the number of $L$ to stabilize training. Second, it removes duplicated positive codes and allows negative sampling to expand the scope of $L$, enabling higher sample efficiency. Finally, $|L|$ becomes a hyperparameter to be controlled.

During training, we use a sampler to keep only one ICD version in each batch, i.e., $L$ is either $L^{icd9}$ or $L^{icd10}$. These batches of different ICD versions are mixed and shuffled to train the model with standard mini-batch gradient descent.  

\subsection{Hyperparameter setting}

We set the dimensions of the encoders and the projection layers the same $d_{note}=d_{code}=d_{shared}$ without further tuning. We follow the implementation from \cite{Edin2023-zj} as best practice in setting the hyperparameters, where \duallaat has a dimension of 512, one bidirectional layer, and dropout rate of 0.3; and \duallaat[cnn] has a 256 filters with filter width of 10, and dropout rate of 0.2. 
All models are trained using a linear scheduler with learning rate of 0.001 and 2000 warmup steps on batches of 32 samples. \duallaat is additionally imposed with weight decay of 0.001 to regularize training.
The maximum number of tokens in clinical notes is set to 4000, and that for the clinical code descriptions is set to 48.

We pretrain the word embeddings (100 dimensions) based on the three train sets, again following the setup in \cite{Edin2023-zj}. By default, we train the models on full ICD codes, and set the label space size $|L|$ to 8192.
We use GRU as the default RNN choice for \duallaat[rnn], with an alternative \duallaat[cnn] using CNN.

\begin{table*}[h!]
\centering

\resizebox{.85\textwidth}{!}{%
\begin{tabular}{lccccc}
    \toprule
    & CAML & LAAT & PLM-ICD & \duallaat[cnn] & \duallaat \\
    & & & & (+ ICD-9) & (+ ICD-9) \\
    \midrule
    AUC-ROC Micro & 98.5 $\pm$ 0.0 & 99.0 $\pm$ 0.1 & 99.2 $\pm$ 0.0 & 99.3 $\pm$ 0.0 & {\bfseries 99.3 $\pm$ 0.0} \\
    AUC-ROC Macro & 91.1 $\pm$ 0.1 & 95.4 $\pm$ 0.3 & 96.6 $\pm$ 0.2 & 97.2 $\pm$ 0.0 & {\bfseries 97.4 $\pm$ 0.0} \\
    F1 Micro      & 55.4 $\pm$ 0.2 & 57.9 $\pm$ 0.1 & {\bfseries 58.5 $\pm$ 0.7} & 56.6 $\pm$ 0.2 & 58.0 $\pm$ 0.0 \\
    F1 Macro      & 16.0 $\pm$ 0.3 & 20.3 $\pm$ 0.4 & 21.1 $\pm$ 2.3 & 26.0 $\pm$ 0.2 & {\bfseries 29.9 $\pm$ 0.3} \\
    P@8           & 66.8 $\pm$ 0.2 & 68.9 $\pm$ 0.1 & {\bfseries 69.9 $\pm$ 0.6} & 68.4 $\pm$ 0.2 & {\bfseries 69.9 $\pm$ 0.1} \\
    P@15          & 52.2 $\pm$ 0.1 & 54.3 $\pm$ 0.1 & {\bfseries 55.0 $\pm$ 0.6} & 53.6 $\pm$ 0.2 & 54.9 $\pm$ 0.1 \\
    R-precision   & 54.5 $\pm$ 0.2 & 57.2 $\pm$ 0.1 & 57.9 $\pm$ 0.8 & 56.4 $\pm$ 0.2 & {\bfseries 58.0 $\pm$ 0.1} \\
    MAP           & 57.4 $\pm$ 0.2 & 60.6 $\pm$ 0.2 & 61.9 $\pm$ 0.9 & 60.4 $\pm$ 0.2 & {\bfseries 62.3 $\pm$ 0.1} \\
    \bottomrule
\end{tabular}
}
\caption{Results on frequent ICD-10 codes (main benchmark) with additional baseline models. Scores are reported with standard deviation.}
\label{tab:std}
\end{table*}

\end{document}